\documentclass[journal]{IEEEtran}

\usepackage{framed,multirow}

\usepackage{amssymb}
\usepackage{latexsym}

\usepackage{url}
\usepackage{xcolor}
\definecolor{newcolor}{rgb}{.8,.349,.1}

\usepackage{times}
\usepackage{epsfig}
\usepackage{graphicx}
\usepackage{amsmath}
\usepackage{amssymb}

\usepackage{float}
\usepackage{stfloats}
\usepackage{multirow}
\usepackage{booktabs}
\usepackage{tablefootnote}
\usepackage{makeidx}
\usepackage[switch]{lineno}
\usepackage{color}

\newcommand{\eg}{e.g.}
\newcommand{\ie}{i.e.}
\newcommand{\etc}{etc.}

\begin{document}
	\title{Learning deep forest with multi-scale Local Binary Pattern features for face anti-spoofing}
	
	\author{Rizhao~{Cai},
		Changsheng~{Chen},~\IEEEmembership{Member,~IEEE},
		
		\thanks{Rizhao Cai is now with the Rapid-Rich Object Search
			(ROSE) Lab, Nanyang Technological University, Singapore. This work was done when he was an undergraduate student at Shenzhen University, China (email: cairizhao@email.szu.edu.cn).}

		\thanks{Changsheng Chen is currently with the Guangdong Key Laboratory of Intelligent Information Processing and Key Laboratory of Media Security,College of Electronics and Information Engineering, Shenzhen University, Shenzhen, China, and also with Shenzhen Institute of Artificial Intelligence and Robotics for Society, China (e-mail: cschen@szu.edu.cn).}
	
	}	
	\markboth{{Submitted to \textit{Pattern Recognition Letter} in {December} 2018}}%
	{Shell \MakeLowercase{\textit{et al.}}: Bare Demo of IEEEtran.cls for IEEE Journals}

	\maketitle

		\begin{abstract}
			    Face Anti-Spoofing (FAS) is significant for the security of face recognition systems. Convolutional Neural Networks (CNNs) have been introduced to the field of the FAS and have achieved competitive performance. However, CNN-based methods are vulnerable to the adversarial attack. Attackers could generate adversarial-spoofing examples to circumvent a CNN-based face liveness detector.  Studies about the transferability of the adversarial attack reveal that utilizing handcrafted feature-based methods could improve security in a system-level. Therefore, handcrafted feature-based methods are worth our exploration. In this paper, we introduce the deep forest, which is proposed as an alternative towards CNNs by Zhou et al., in the problem of the FAS. To the best of our knowledge, this is the first attempt at exploiting the deep forest in the problem of FAS. Moreover, we propose to re-devise the representation constructing by using LBP descriptors rather than the Grained-Scanning Mechanism in the original scheme. Our method achieves competitive results. On the benchmark database IDIAP REPLAY-ATTACK, 0\% Equal Error Rate (EER) is achieved. This work provides a competitive option in a fusing scheme for improving system-level security and offers important ideas to those who want to explore methods besides CNNs.
			\\
		\end{abstract}
		

	
	\section{Introduction}	
	Face recognition systems, which identify an individual with her/his face, have been widely used in practical applications such as mobile phone unlocking. However, the existing face recognition techniques cannot differentiate between genuine faces (captured from human) and spoofing faces (captured from the faces in images, digital display, \etc). Most of the face recognition systems are therefore vulnerable to Presentation Attack (PA), including print attack, replay attack. Attackers could bypass the face recognition systems by presenting different types of spoofing faces since face images can be readily available to attackers from social platforms, \eg, Facebook, Instagram  \cite{patel2016secure}. To guarantee the security of face recognition systems, there are increasing demands for developing the FAS techniques.
	\begin{figure}[t]
		\centering
		\includegraphics[width=0.85\linewidth]{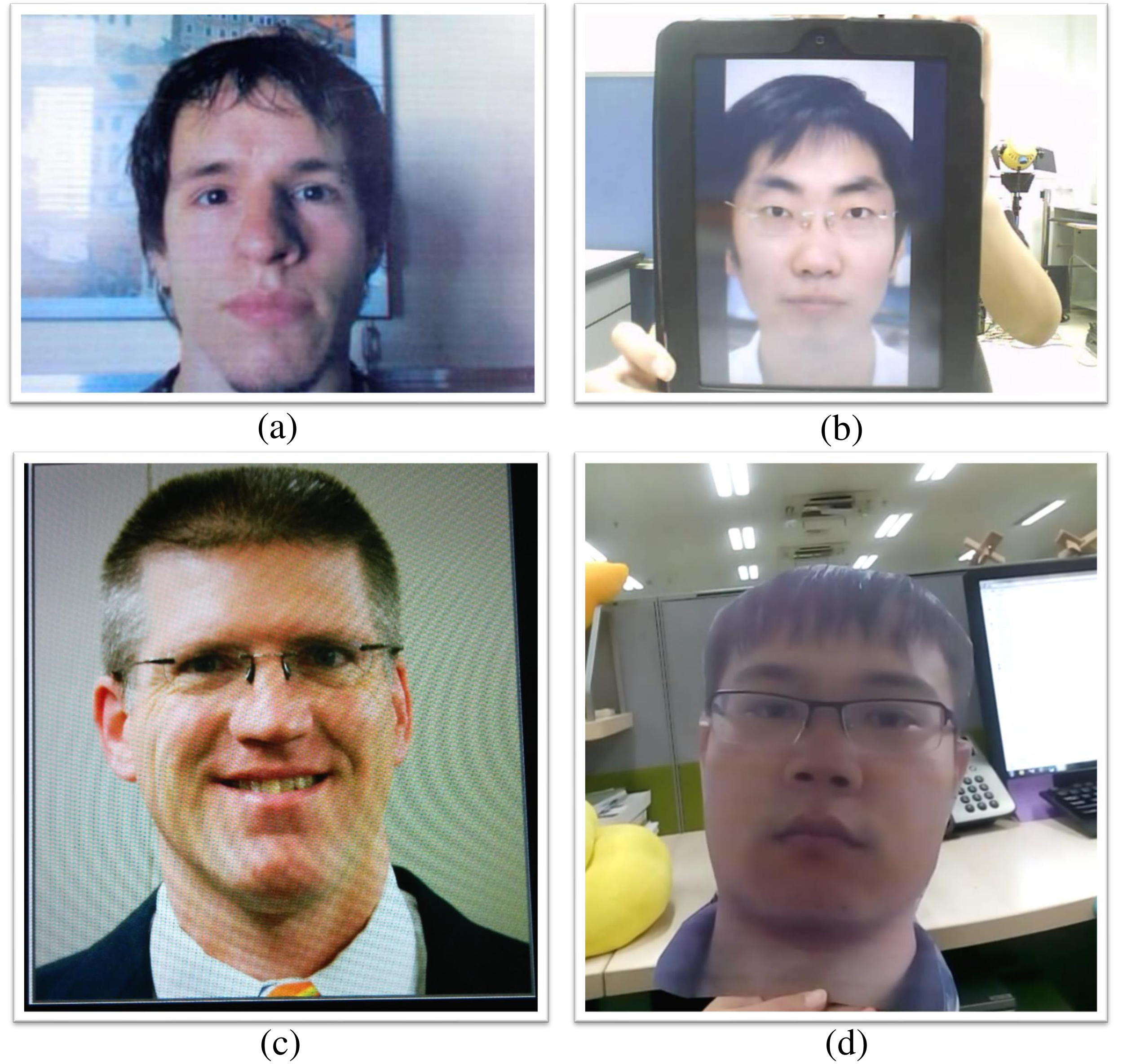}
		\caption{Examples of Presentation Attack (PA). (a): display attack. The face is in a digital display screen \cite{replayattack}. (b): replay attack \cite{CASIAFASD}. The face is in a video. (c): print attack. The face is in a print photo \cite{patel2016secure}. (d): print attack. The face is in a print photo that is tailored \cite{ROSE}. }\label{fig:face}
		\hspace{-5cm}
	\end{figure}	

	\par  Traditionally, image descriptors, such as Local Binary Pattern (LBP) and Scale Invariant Feature Transform (SIFT), are utilized to extract features for describing the data from the FAS databases. Recently, with the powerful ability for learning deep representations from data, Convolutional Neural Networks (CNNs) have been successfully exploited in various visual tasks, \eg objects classification \cite{krizhevsky2012imagenet}, face recognition \cite{zhu2013deep}, \etc and have achieved the state-of-the-art performances. The attempts of CNNs in the FAS have been also reported and have achieved much improvement \cite{Yang2014Learn,krizhevsky2012imagenet,menotti2015deep, Xu2016Learning,sigportLSTMface}. 
	
	

	    Although CNN-based methods have shown their excellent capacities,  it is pointed out that they are vulnerable to adversarial attack \cite{adv_GoodFellow_2016_CoRR_intriguing}. Under such adversarial attack, a CNN-model would fail to correctly classify the adversarial examples, which are generated by imposing some human-invisible perturbations on the original samples. What is more, though adversarial examples are usually manipulated in the digital world, they could still take effects even after a print-and-capture cycle \cite{adv_GoodFellow_2016_CoRR_PhysicalAttack, adv_DawnSong_2018_CVPR_AttackSign, adv_Sharif_2016_CCS_glassFace}. In other words, the adversarial attack can be conducted in the physical world. Worse still, the adversarial examples are shown to be transferable. Empirical experiments in \cite{adv_DawnSong_2017_ICLR_Transfer, securekernel2016} and theoretical analysis in \cite{transfer_space} show that adversarial examples can be transferred to attack other models as long as they adopt the same or similar features even if the classification models are different (Support Vector Machine, Random Forest, \etc). Therefore, it is likely for attackers to generate adversarial-spoofing examples to attack a CNN model for face liveness detection in a face recognition system.	
		
		Fortunately, using handcrafted feature-based methods could be a solution. In \cite{securekernel2016, transfer_space}, it is revealed that adversarial examples are non-transferable when they are in the different feature spaces as the input of their victim models. This indicates that the handcrafted features from RGB images as input for a face anti-spoofing model could be an approach against the adversarial-spoofing attack targeted at the CNN-based models. In the cybersecurity applications, it is also suggested in \cite{DBLP} that ensembling a diverse pool of models of different features could improve the security of a cyber system against the adversarial attack. Hence, to alleviate the threats of the adversarial attack, handcrafted feature-based methods also deserve efforts of exploration.
	
	\par In this paper, we introduce a new feature-based method, the deep forest \cite{gcforest}, to the FAS problem. The deep forest is an advanced synthesis of tree-ensemble methods. It consists of the Grained-Scanning Mechanism (GSM) for learning representations from data and the layer-cascade strategy for further processing the representations. The deep forest has been evaluated on several visual tasks, \eg, face recognition, handwriting recognition, \etc, and it achieves competitive performance \cite{gcforest}. Since the deep forest is newly-published, there are not yet many works about using the deep forest in applications related to biometrics. To the best of our knowledge, we are the first to introduce the deep forest in the problem of the FAS. However, the performance is not satisfactory in our initial attempt when the GSM, proposed by \cite{gcforest}, is directly used to learn representations for the spoofing detection. The unsatisfactory result suggests that the GSM is not competent enough in capturing the cues for the face spoofing detection. Inspired by texture analysis \cite{maatta2011face, Yang2013Face, Nosaka2011Feature}, the baseline approaches in the research area of the FAS, we propose to employ Local Binary Pattern (LBP) descriptors to construct the representations of spoofing information. Experimental results show that the proposed approach has achieved competitive performance. 
	
	\par $\bullet$ To the best of our knowledge, this is the first work that introduces the deep forest to the problem of FAS. Our method offers an important reference and a competitive option to those who want to fuse diverse methods in their schemes for system-level security in their cases.
	\par $\bullet$ We re-devise the representation constructing by utilizing the LBP descriptors instead of the GSM. The proposed scheme that integrates LBP descriptors and the deep forest learning method achieves better results than that of the GSM \cite{gcforest}.
	\par $\bullet$ The proposed scheme shows competitive performance compared to state-of-the-art approaches. On the IDIAP REPLAY-ATTACK database \cite{replayattack}, 0\% Equal Error Rate (EER) is achieved. Also, extensive experiments on the two newly-published databases, MSU USSA database and ROSE-YOUTU database have been conducted. On the MSU database, EER of 1.56\% is obtained, which is a competitive result compared to the Patch-based CNN (0.55\% EER) and the Depth-based CNN (2.62\% EER) proposed by \cite{Atoum2018Face}.

	\par The rest of the paper is organized as follow: Section \ref{sec: review} presents brief literature reviews about approaches to FAS and about learning methods that are forest-related. The proposed scheme is elaborated in Section \ref{sec:method}. The performance of the proposed scheme is evaluated in Section \ref{sec:exp}. Finally, Section \ref{sec:end} concludes this paper.
	\section{Related Works} \label{sec: review}
	
	\begin{figure}[t]
		
		\begin{center}
			\includegraphics[width=0.85\linewidth]{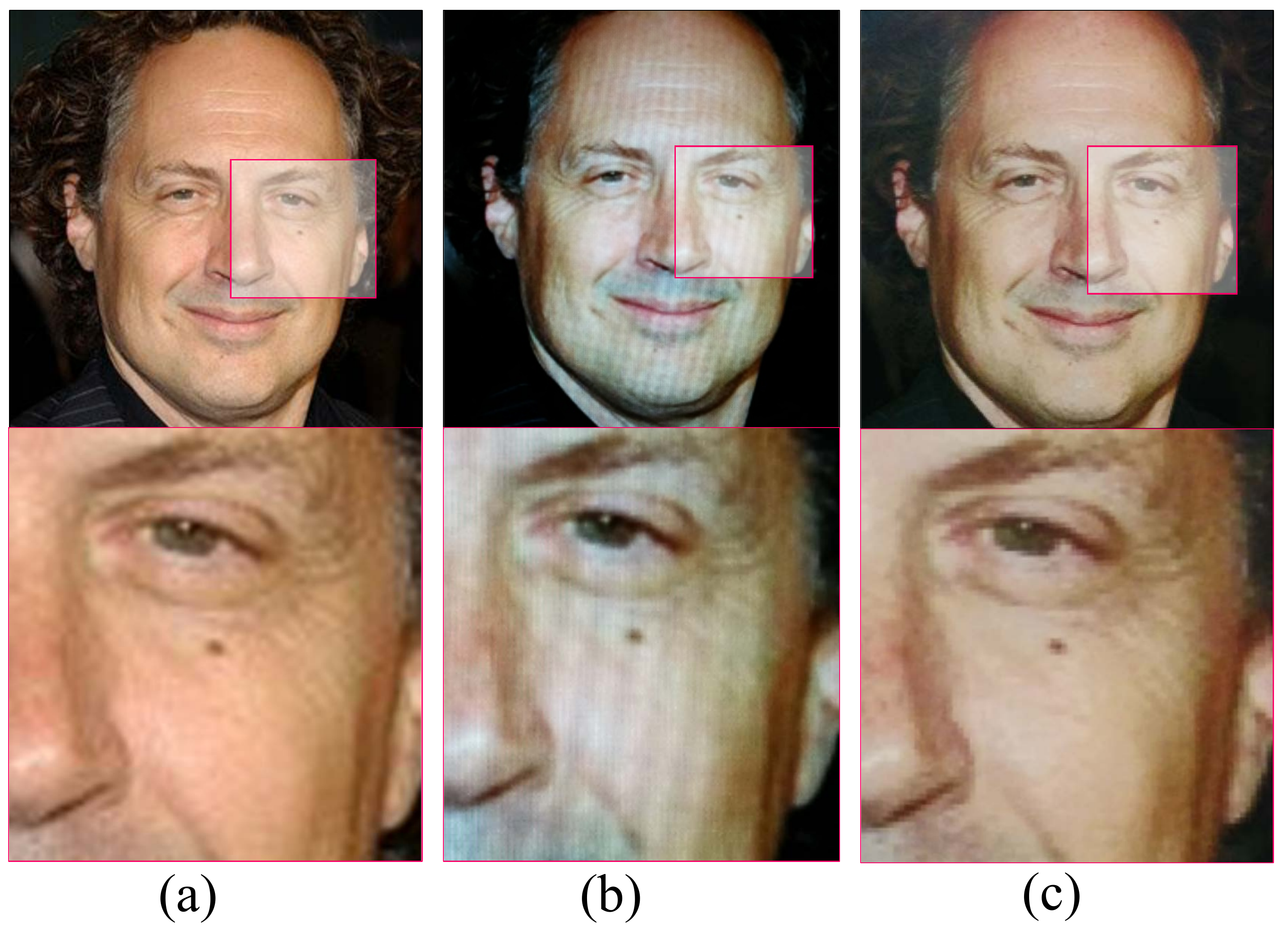}
			\caption{Examples of genuine faces and spoofing faces from MSU USSA database \cite{patel2016secure}. Columns (a), (b), (c) are the genuine face, display face and printed photo face and their correspondent magnifying regions, respectively.}\label{fig:texture}
		\end{center}
		\hspace{-20cm}
	\end{figure}
	
	In this section, the literature on both traditional handcrafted feature-based methods and CNN-based methods in the problem of FAS is first reviewed, followed by the tree-ensemble learning methods.
	\subsection{The Existing Works on Face Anti-Spoofing}
	\subsubsection{The Traditional Methods}
	Most of the traditional FAS approaches focus on designing handcrafted features and learning classifiers with traditional learning methods, e.g., Support Vector Machine (SVM) \cite{Atoum2018Face}. Texture analysis is one of the main approaches to spoofing faces detection since there are inherent texture disparities between genuine faces and spoofing faces of the print attack or of the replay attack. As can be seen in Fig.~\ref{fig:texture}, images of the spoofing faces, compared to the genuine faces, usually have lower quality and contain visual artifacts because of the recapturing process. These disparities can be described effectively by texture descriptors. Relevant methods aimed at capturing these disparities in the Fourier spectrum or spatial domain are reported. Ref.\cite{Tan2010Face} uses Difference-of-Gaussian (DoG) features to describe the disturbance of frequency resulting from the recapturing. Besides, the Local Phase Quantization (LPQ) that analyzes distortion through the phase is also discussed by \cite{Gragnaniello2015An}. In addition, in the spatial domain, a significant number of research works employ the LBP-based features to describe the disparities from local texture information \cite{maatta2011face, Yang2013Face, Nosaka2011Feature}. Analogously, methods that utilize Scale-Invariant Feature Transform (SIFT) and Speed-Up Robust Feature \cite{Boulkenafet2017Face} are also reported. Besides,  to utilize motion information from the temporal domain, the texture-based methods mentioned above are extended into three orthogonal planes, \eg, LBP-TOP \cite{Pereira2012LBP}, and LPQ-TOP \cite{Arashloo2017Face}. Moreover, the color information of spoofing faces, which is less abundant after distortions in the recapturing process, is essential in discriminating spoofing faces. Therefore, color texture methods are proposed in \cite{Color2017Face} by extracting features from separate channels in a certain color space (\eg, to extract features of images in HSV space from the three components H, S, and V individually) using the aforementioned methods. 
	\begin{figure}[t!]		
		\centering	
		\includegraphics[width=0.9\linewidth]{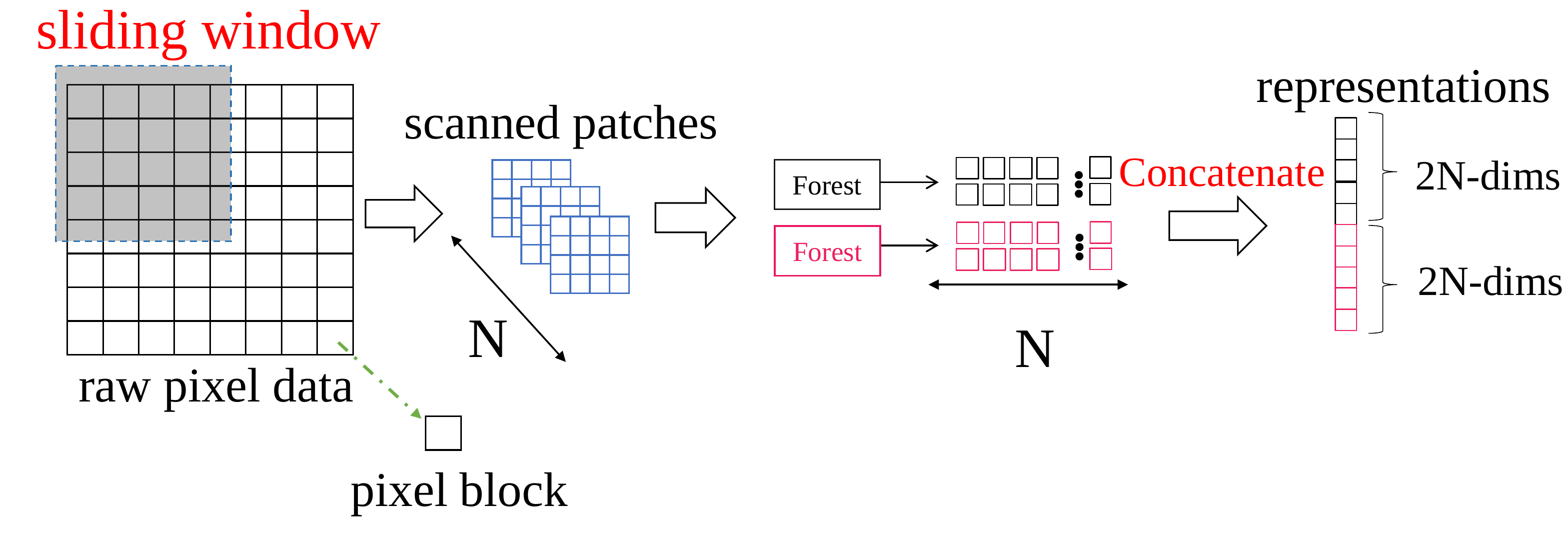}
		\caption{The illustration of how the GSM learns representations for local information \cite{gcforest}. First, a sliding window with a certain stride is used to scan raw pixels. Then, all the scanned patches are fed to forests, a random forest (black) and a completely-random forest (rose). Finally, all the output results from the forests will be concatenated as the representations of the raw pixel data. For full details about the GSM, please refer to \cite{gcforest}}. \label{fig:scanning}
		\hspace{-20cm}
	\end{figure}  
	\subsubsection{The Deep-learning Based Methods}
	Recently, CNN-based methods with the powerful ability for learning deep representations from data has attracted many research attention. Ref.\cite{Yang2014Learn} trains a CNN to learn deep representations for face anti-spoofing based on the AlexNet architecture \cite{krizhevsky2012imagenet}. After that, the feasibility of CNN in learning deep representation for biometric, including face anti-spoofing, is further demonstrated by \cite{menotti2015deep}, and more CNN-based methods are increasingly reported \cite{Atoum2018Face, Haoliang2}. In addition, efforts in exploiting Long Short-Term Memory networks (LSTM) to utilize temporal information from frames of videos are also reported in \cite{Xu2016Learning,sigportLSTMface}.

	\subsection{The Tree-Ensemble Methods}

	\begin{figure}[t!]
		\centering
		\includegraphics[width=0.958\linewidth]{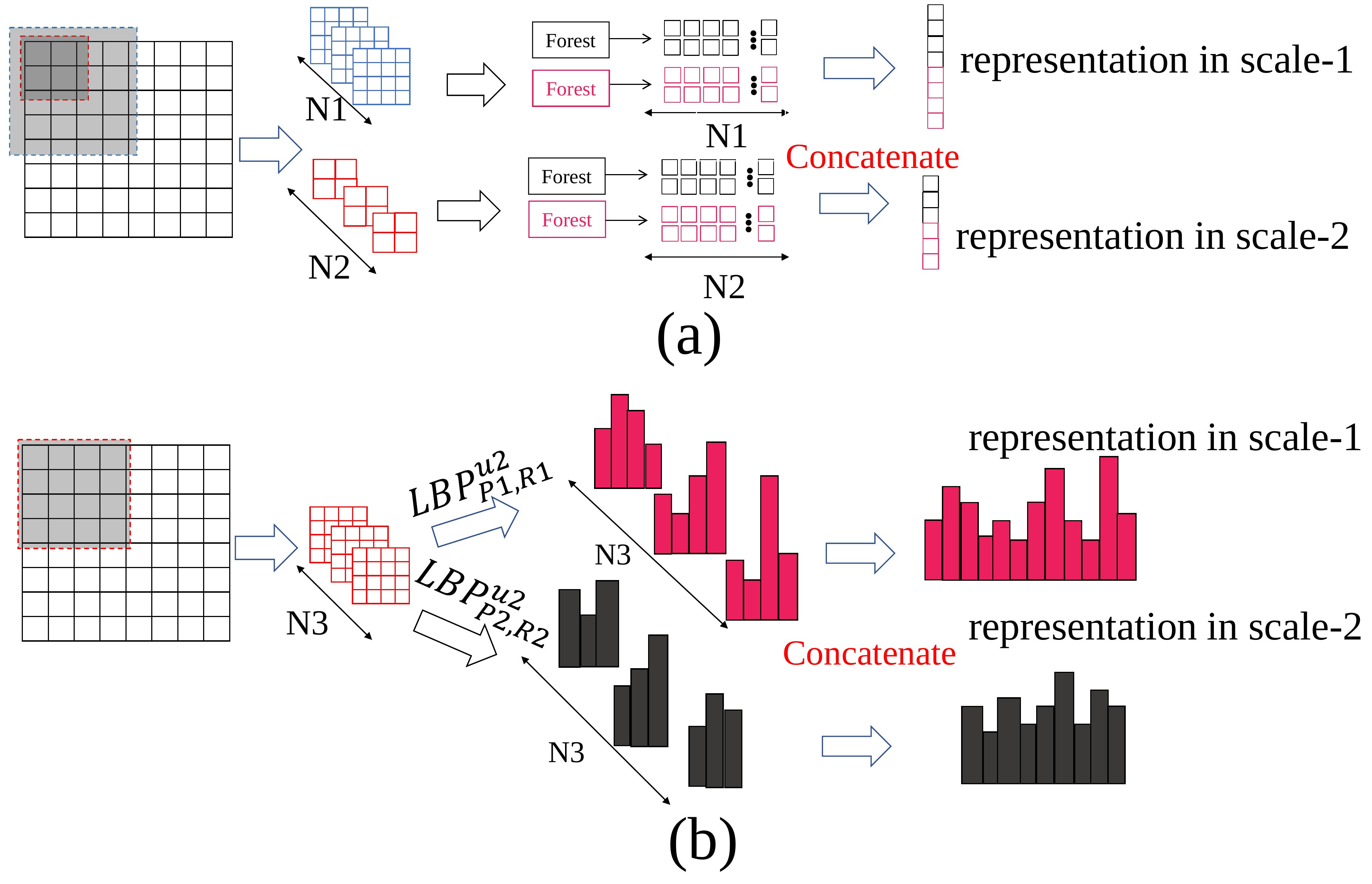}
		\caption{Illustrations of constructing multi-scale representations. The (a) and (b) illustrate how the MGSM and the proposed scheme construct representations on multi-scales respectively.}\label{fig:multiscale}
	\end{figure}	
	The tree-ensemble methods are based on decision trees. The random decision forest is first proposed as a solution towards the dilemma between performance and generalization of the decision tree \cite{Ho1998The}. It is later ameliorated to the Random Forest (RF) by introducing the feature sampling and the data bootstrapping by \cite{Breiman2001Random}. Completely-Random tree Forest (CRF) has a mechanism that is much more ``random'' than RF since it splits the nodes randomly, regardless any criterions \cite{Liu2008Spectrum}.
	Both the RF and the CRF would project original features into subspaces by sampling the original features. This reduces dimensions of features to process, which facilitates the handling of high dimensional features \cite{gcforest}. The Gradient Boosting Decision Tree (GDBT) methods introduce loss functions for training which have not been included in the RF and the CRF. The GDBT models are trained by boosting the gradients of the loss. An effective way to implement GDBT is proposed by \cite{xgboost}, namely the XGBoost. The XGBoost provides a more flexible and powerful scheme that approximates non-differentiable loss functions by the first two terms of their Taylor Expansion, so users are enabled to define arbitrary loss functions in their problems. The XGBoost has achieved superior performance among many GDBT implementations.
	\par The deep forest, proposed by \cite{gcforest}, can achieve state-of-the-art performance compared to CNN-based methods on several visual tasks reported by \cite{gcforest}. It is proposed in \cite{gcforest} that a Grain Scanning Mechanism (GSM) is used to learn representation from data and a cascade strategy for further processing the representations. The RF is a basis of the GSM of the deep forest and CRF offers another option for the deep forest. By combining different types of forest, the diversity of representations learned by the deep forest can be improved \cite{gcforest}. The XGBoost and other implementations of GBDT can also be a basis in the deep forest. More details about the deep forest can be found in \cite{gcforest}. Unlike CNNs, whose structures are fixed during the training process, the number of cascade levels of the deep forest model depends on the scale of the data and grows as the training proceed. Once the output scores (accuracy, loss, \etc) begins to converge, the growth stops. Hence, the complexity of the model can be adaptively adjusted according to the scale of the database. This ensures that the deep forest can maintain a satisfactory result even on a small-scale database \cite{gcforest}.

	\section{The Proposed Deep Forest with LBP Features} \label{sec:method}
	\par The LBP is selected due to two reasons. Firstly, LBP features cannot be reconstructed back to RGB pixels images, thus helpful against the adversarial-spoofing attack. Secondly, LBP is designed for texture description, which may be appropriate for the face spoofing problem. This section will first elaborate on how to use the LBP descriptors \cite{menotti2015deep} to leverage texture information. Then, the proposed scheme integrating the deep forest and the LBP features will be presented.
	
	\begin{figure*}[t]
		\centering
		\includegraphics[width=0.8\linewidth]{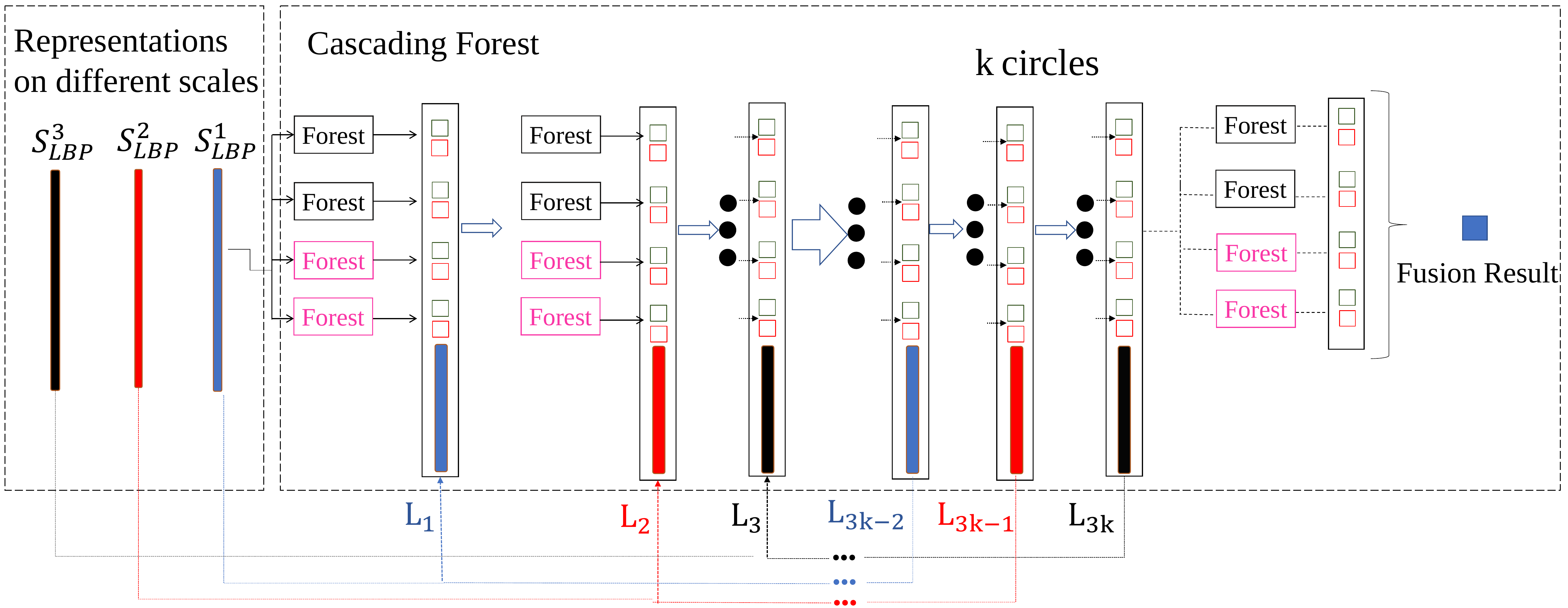}
		\caption{The procedure of the deep forest learning with multi-scale representations. The left part contains the LBP representations on three different scales, denoted by $S_{LBP}^1$, $S_{LBP}^2$ and $S_{LBP}^3$, respectively. The right part illustrates the cascading strategy. The ``black" and ``red" boxes are output results of each forest from the previous layer. They will be concatenated with features on different scales in different layers as the input of their next layers}\label{fig:architecture}
		\hspace{-5cm}
	\end{figure*}	
	
	\subsection{The LBP-based Features for Texture Analysis}
	\par The Local Binary Pattern (LBP) descriptor proposed by \cite{Ojala2002Multiresolution} is a grey-scale descriptor that is effective for texture description. By calculating the LBP values of the binary patterns for each pixel and accumulating the occurrences of them into histograms, the LBP features can be extracted to represent local texture information. The calculation of LBP can be described as
	\[LBP_{P, R} = \sum_{n=1}^P {\rm sgn} (r_n-r_c) \times 2^{n-1} \tag{1}\]
	where ${\rm sgn}(\cdot)$ takes the sign of the operand, $r_c$ denotes the intensity value of the central pixel and $r_n (n=1, 2,..., P)$ denotes the intensity values of $P$ adjacent pixels distributed symmetrically at a circle of radius $R (R > 0)$. An image can be divided into several patches, and LBP histograms are calculated for each patch. Then, all the histograms can be concatenated into a feature vector to represent the image in the texture field. To fully exploit the color information, color LBP features will be employed by referring to \cite{Color2017Face} in this paper. The color LBP features are to extract LBP features from each component individually of the color space (\eg Red, Blue, Green in the RGB space or Hue, Saturation, Value in the HSV space) and the obtained results will be concatenated into a feature vector \cite{Color2017Face}. These features based on LBP descriptors are to be called LBP features in this paper.
	
	\par The GSM learns the representations of local information from adjacent pixels within a certain window, and similarly, the extraction of LBP-based features also considers the local information. On the other hand, the significant contrast between employing the GSM~\cite{gcforest} (illustrated in Fig.~\ref{fig:scanning}) and the LBP features lies in the representations constructing. The GSM constructs representations by learning from data while the LBP features construct representations with the domain knowledge of a researcher.

	\subsection{The Proposed Multi-scale Representations}	
	Firstly, we propose to use the multi-scale LBP descriptor to construct the multi-scale representations. Taking multi-scales into accounts is important because the image samples are from practical capturing conditions and there are variations of the textural disparities. For example, although both Fig.~\ref{fig:texture}~(b) and (c) are spoofing faces, they are captured under different conditions, \ie, different devices, different circumstances, \etc, so they show different texture appearances in both patterns and scales. Therefore, different scales of local information should be taken into considerations.
	As is illustrated in Fig.~\ref{fig:multiscale}~(a), the Multi-Grained Scanning Mechanism (MGSM) \cite{gcforest} is used to learn representations from data on multiple scales. By changing the size of the sliding windows and conducting the GSM, relationships of the pixels on different scales will be learned, and local information on different scales can be leveraged \cite{gcforest}. On the other hand, Fig.~\ref{fig:multiscale}~(b) illustrates our proposed scheme. In the proposed scheme, there is a sliding window for scanning patches of pixels, and ${\rm LBP}_{P, R}^{u2}$ descriptors \cite{Ojala2002Multiresolution} are used to obtain LBP features. By changing the parameters $P$ and $R$, representations on different scales can be obtained. To utilize color information, the color LBP features will be adopted in the proposed scheme to construct representations on different color channels and scales according to \cite{Color2017Face}. One of the differences between the MGSM and our proposed scheme in constructing multi-scale representations lies in the selection of sliding windows. With the MGSM, windows of different sizes are needed to learn multi-scale representations, while multi-scale representations based on LBP features can be obtained with a fixed size window. This is because the representations on a certain scale learned by the GSM only depends on the size of the window; while, in the exploitations of LBP descriptors, the representation in a certain scale can also be determined by certain parameters of LBP descriptors, \ie, $P$ and $R$. Multiple sizes of windows are not adopted in this paper for the consideration that when small-size windows are used to extract LBP histograms, many of the bins are empty, and the obtained features will be of high-dimension and sparse, i.e., less informative.
	
	\par Secondly, instead of concatenating all the representations on these three scales to construct a feature vector, as performed in some traditional methods \cite{Color2017Face, patel2016secure}, a circular cascading strategy is adopted in our proposed scheme by referring to \cite{gcforest}. This strategy is shown in Fig.~\ref{fig:architecture}. The $n$-th layer will be identified as ${\rm L}_n$. Representations on the three scales are denoted by $S_{LBP}^1$, $S_{LBP}^2$ and $S_{LBP}^3$, respectively. They will be individually fused with the output of each layer of the deep forest, and each layer will focus on the representations on a certain scale. The $S_{LBP}^1$ will be fed to the first layer of deep forest and fused with the output of ${\rm L_1}$. Then, the representation $S_{LBP}^2$ will be fused with the output from ${\rm L_1}$ and become the input of ${\rm L_2}$. The $S_{LBP}^3$ and ${\rm L_3}$ will do so. It should be noted that this cascading process is circular. For instance, in the next circle, the $S_{LBP}^1$ is concatenated in the ${\rm L_4}$. In the $k$-th circle, the $S_{LBP}^1$ will be concatenated in the ${\rm L}_{3k-2}, k\in \mathbb{N}$. Actually, the options of the scales and cascade strategies are flexible according to tasks.
	
	\section{Experiments}\label{sec:exp}		
	In this section, a brief introduction for four databases, on which the experiments are conducted, will be first given. Then, the details about the settings of the experiments are shown. Finally, the experimental results are presented and discussed.
	\subsection{Databases}
	In our experiment, several representative databases have been employed. Two are benchmark databases, CASIA FASD \cite{CASIAFASD} and IDIAP REPLAY-ATTACK \cite{replayattack}, and two newly-published databases, ROSE-YOUTU LIVENESS database \cite{ROSE} and MSU USSA database \cite{patel2016secure}. The IDIAP, CASIA and ROSE-YOUTU databases consist of videos, covering replay attack, display attack, and print attack. The MSU database only contains images, \ie, only including display attack and print attack. More specifically, the scales of each database is summarized below.
	\par The IDIAP REPLAY-ATTACK database \cite{replayattack} constitutes about 50 subjects. There are 60 videos of genuine faces and 300 videos of fakes faces in the training set. In the testing set, there are 80 videos of genuine faces and 400 videos of spoofing faces.
	\par The CASIA database \cite{CASIAFASD} consists of 600 videos from 50 subjects, 20 subjects for the training set and 30 subjects for the testing set. For each subject, there are 3 videos of genuine faces and 9 videos of spoofing faces.
	\par The ROSE-YOUTU LIVENESS database \cite{ROSE} contains 10 and 12 subjects in the training set and the testing set, respectively. For each subject, there are 180 videos, consisting of various types of attack and light conditions. This database is the latest database concerning PA, and there is a tailored print attack. As can be seen in Fig.~\ref{fig:face}, the background is not included in the recapturing process, making this database more challenging.
	\par The MSU USSA database \cite{patel2016secure} includes 1,000 genuine faces (about 1,000 subjects) and about 6,000 spoof face images. There is no division of the training set and the testing set, so a 5-fold validation protocol is used to evaluate the performance of the FAS methods.
	
	\subsection{Experimental Setups}\label{subsec:Preprocessing}
	\par In the first place, it should be highlighted that some data preprocessing have been performed in our experiments. When conducting experiments on the MSU and IDIAP databases, the whole image frames are taken as the inputs for making full use of information. This is because the PA places the spoof media near the cameras to achieve high recapturing quality, and the ``background'' is also recaptured (as shown in (a) and (c) of Fig.~\ref{fig:face}). The recaptured background provides useful information which is beneficial to the spoof face detection. However, in the CASIA database and the ROSE-YOUTU database, the PA is far away from cameras and hence the background is not included in the recapturing process (as shown in (b) and (d) of Fig.~\ref{fig:face}). Under this circumstance, if the whole frame is used as the input, unnecessary interference (from the genuine background) will be introduced. Therefore, the Viola-Jones method \cite{viola} is utilized to detect the faces in frames from the ROSE-YOUTU and CASIA databases. The detected faces are cropped and are employed as the inputs in the experiments. Then the resolution of all the inputs (i.e., whole frames from the IDIAP databases and MSU database or cropped face regions from the CASIA database and ROSE-YOUTU database) is normalized to $128 \times 128$ pixels for a trade-off between the computational complexity and performance by referring to the prior works \cite{patel2016secure}.
	\par Secondly, the settings of the experiments are elaborated. In \cite{gcforest}, three square sliding windows of different sizes are employed to evaluate the performance of the deep forest. By referring to this, three scales of windows, 16, 32 and 64 pixels, with the strides of 8, 16, and 32 pixels, respectively, are used for the MGSM in this paper. The obtained representations on these three scales will be denoted by $S_{GSM}^1$, $S_{GSM}^2$, $S_{GSM}^3$, respectively. In our proposed scheme, the size of the sliding window is fixed as 32 pixels and the stride is 16 pixels. For each image of $128 \times 128$, there will be $7 \times 7$ overlapped sub-patches in total. Three ${\rm LBP}_{P, R}^{u2}$ descriptors \cite{Ojala2002Multiresolution}, ${\rm LBP}_{8,1}^{u2}$, ${\rm LBP}_{16,2}^{u2}$ and ${\rm LBP}_{24,3}^{u2}$, are utilized to construct representations on three scales, and the obtained representations are referred to as $S_{LBP}^1$, $S_{LBP}^2$, $S_{LBP}^3$, respectively. Also, color LBP features in HSV and YCbCr spaces are considered in this paper. That is to extract features in each seperate channels of a image. For one patch, the feature lengths of color (RGB, HSV, YCbCr) ${\rm LBP}_{8,1}^{u2}$, ${\rm LBP}_{16,2}^{u2}$ and ${\rm LBP}_{24,3}^{u2}$ are $59 \times 3$, $59 \times 3$, $59 \times 3$, respectively. Since there are 49 sub-patches for each image, the lengths of the final $S_{LBP}^1$, $S_{LBP}^2$, $S_{LBP}^3$ will be $49 \times 59 \times 3$, $49 \times 243 \times 3$, $49 \times 555 \times 3$ respectively. During the cascading operation, $S_{LBP}^1$ / $S_{GSM}^1$ will be fused with ${\rm L_1}$, $S_{LBP}^2$ / $S_{GSM}^2$ with ${\rm L_2}$ and $S_{LBP}^3$ / $S_{GSM}^3$ with ${\rm L_3}$. This process continues circularly until the training process terminates. This process will stop automatically when accuracies converge for several rounds. As for the setting of forests utilized in the deep forest, four RFs and four CRF are employed and there are 500 trees in each forest by referring to \cite{gcforest}. These are implemented with the package of the gcForest\footnote{https://github.com/kingfengji/gcForest} with default settings of the forests. For more details of the mechanism about the deep forest, please refer to \cite{gcforest}.
	
	\subsection{Experimental Results} \label{sec:expresult}
	
	\subsubsection{Comparisons between Multi-scale Representations} \label{exp-1}
	\begin{table}[tbp]
		\footnotesize
		\begin{center}
			\caption{Comparisons between two implementations of multi-scale representations on MSU USSA database, IDIAP database, CASIA database. Performance is evaluated by EER (\%).}
			\begin{tabular}{|l|c|c|c|}
				\hline
				\multicolumn{1}{|l|}{\multirow{2}[4]{*}{Multi-scale\newline{} representations}} & \multicolumn{3}{c|}{Database} \\\cline{2-4}
				& \multicolumn{1}{c|}{MSU} & \multicolumn{1}{c|}{IDIAP} & \multicolumn{1}{c|}{CASIA} \\
				\hline
				GSM (RGB) \cite{gcforest} & 4.84 & 1.02 & 14.50 \\
				\hline
				proposed (RGB) & 4.17 & 0 & 11.82 \\
				\hline
				proposed (HSV) & 2.14 & 0.052 & 8.73 \\
				\hline
				proposed (YCBCR) & 1.56 & 0 & 9.66 \\
				\hline
			\end{tabular}%
			\label{tab:cmp}
		\end{center}
		\hspace{-5cm}
	\end{table}%
	Table~\ref{tab:cmp} provides the experimental results of the GSM and of the proposed scheme in terms of Equal Error Rate (EER). From Table~\ref{tab:cmp}, by integrating the LBP features (RGB) with the deep forest, the EER on the MSU, IDIAP, CASIA databases are reduced from 4.84\% to 4.17\%, from 1.02\% to 0\% and from 14.50\% to 11.82\%, respectively. These results suggest that LBP-based features are more competent in exploiting texture information to represent the degradation of the spoofing faces than the GSM. Furthermore, across different color spaces, the performances of LBP features in HSV and YCbCr color spaces are generally better than that in the RGB color space. This is because the change of illuminance should not interfere chrominance information, which is crucial in color texture methods, and the HSV space and YCbCr space separate primely components of illumination and chrominance. However, the RGB space remains high correlations in the three components, and slight variance of illumination by altering the R, G, B may result in unexpected change chrominance, making feature less effective \cite{Color2017Face}.

	\begin{figure}[tbp]
		\centering
		\includegraphics[width=0.9\linewidth]{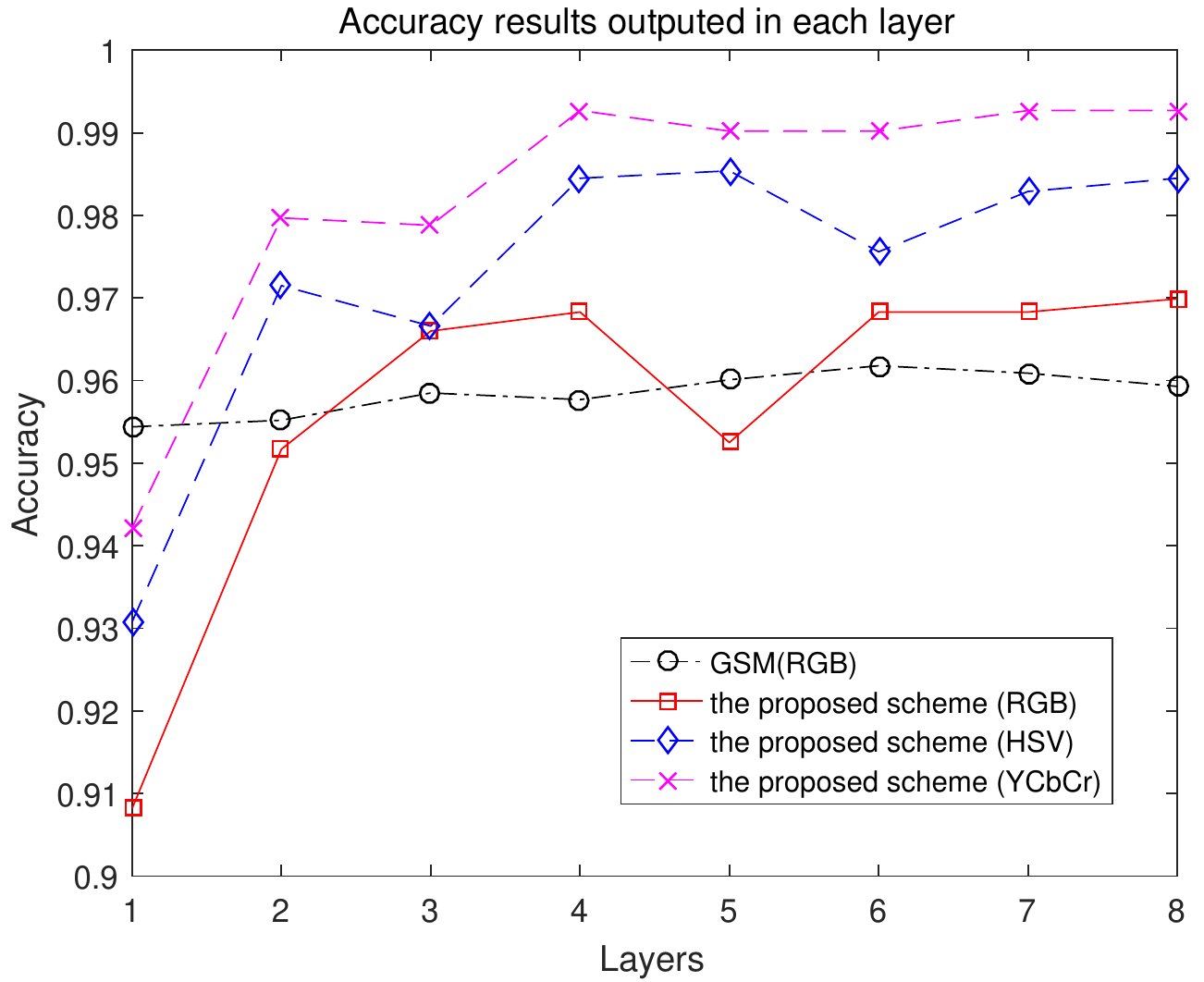}
		\caption{The convergence curves of experiments on MSU database. An average of the results of the five validations is taken. The $x$-axis refers to the number of the cascade layer that increases along with the training process. The $y$-axis refers to the testing accuracy of the output of each layer. }\label{fig:acc}
		\hspace{-5cm}
	\end{figure}

	To further probe into the effectiveness of the proposed scheme, curves of the training accuracy outputted by each layer are drawn and shown in Fig.~\ref{fig:acc}. An upward trend of the accuracy results can be seen. It goes up along with the growth of the structure. There are limited improvements in the curve of over different layers with the GSM, which indicates that the GSM is not able to capture the texture information of the spoofing cues over different scales efficiently. Meanwhile, despite inferior accuracies in the first two layers, the accuracies of the proposed scheme (RGB, HSV, YCbCr) finally outperform that of the GSM. Moreover, the trend of the curves indicates that the cascading strategy enables LBP features to be re-represented. For instance, $S_{LBP}^1$ is fed to the layers ${\rm L_1}$, ${\rm L_4}$ and ${\rm L_7}$, and the outputted accuracies get improved. In layer ${\rm L_1}$, the deep forest model learns from $S_{LBP}^1$, representations on a small scale. Then, after ${\rm L_2}$ and ${\rm L_3}$, where the model has perceived more information from representations on larger scales, the model leads to a better understanding towards the distortion on different scales. 

	\subsubsection{Comparisons with State-of-the-Art Approaches}
	
	\begin{table}[tbp]
		\footnotesize
		\begin{center}
			\caption{Comparisons between the proposed scheme and state-of-the-art approaches on the IDIAP database, CASIA database and ROSE-YOUTU database, which are in terms of EER (\%).}
			\begin{tabular}{|l|c|c|c|}
				\hline
				Method & IDIAP & CASIA & ROSE-YOUTU\\
				\cline{2-4}
				\hline
				LBP-TOP \cite{Pereira2012LBP} & 7.9 & - & - \\
				\hline
				CoALBP (HSV) \cite{Color2017Face} & 3.7 & 5.5 &16.4\\
				\hline
				CoALBP (YCbCr) \cite{Color2017Face}& 1.4 & 10.0 & 17.7\\
				\hline
				Fine-tuned AlexNet \cite{Yang2014Learn}& 6.1 & 7.4 & 8.0\\
				\hline
				CNN+Conv-LSTM \cite{sigportLSTMface}& 5.12 & 22.40 &-\\
				\hline
				CNN+LSTM \cite{sigportLSTMface}& 1.28 & 14.60 &-\\
				\hline
				Patch-based CNN \cite{Atoum2018Face}& 2.5 & 4.44 & - \\
				\hline
				Depth-based CNN \cite{Atoum2018Face}& 0.86 & 2.85 & - \\
				\hline
				proposed (HSV) & 0.052 & 8.73 & 10.9 \\
				\hline
				proposed (YCbCr) & 0 & 9.66 & 11.9\\
				\hline
			\end{tabular}%
			\label{tab:exp-both}%
		\end{center}
		\hspace{-5cm}
	\end{table}%
	Tables~\ref{tab:exp-both} and \ref{tab:exp-msu} provide results of comparisons between the proposed scheme and the state-of-the-art approaches. From Table~\ref{tab:exp-both}, The proposed scheme with simple LBP features is demonstrated to be highly competitive. Firstly, on the CASIA database, the proposed scheme (HSV) achieves 8.73\% EER. Although this result is inferior to results of some CNN-based methods, the patch-based CNN (4.44\%) \cite{Atoum2018Face}, the depth-based CNN (2.85\% ) \cite{Atoum2018Face} and fine-tuned AlexNet (6.1\%) \cite{Yang2014Learn}, it is better than some LSTM-based method with 22.40\% and 14.60\% EERs presented in \cite{sigportLSTMface}. It is worth mentioning that, among the traditional methods (using the SVM classifiers with handcrafted features), particularly among LBP-based methods, Co-occurrence of Adjacent Local Binary Patterns (CoALBP) method \cite{Nosaka2011Feature} has achieved the state-of-the-art performance \cite{ROSE}. Experiments on the CASIA database show that the proposed scheme with LBP features has achieved a better result (9.66\%) than CoALBP (10.0\%) in YCbCr space. Moreover, experimental results on the ROSE-YOUTU database \cite{ROSE}, a more diverse and challenging database, are presented in the last column in Table~\ref{tab:exp-both}. The results show that the CoALBP, which performs well on IDIAP database (3.7\% in HSV and 1.4\% in YCbCr) and CASIA database (5.0\% in HSV and 10.0\% in YCbCr), drops dramatically (16.4\% in HSV and 17.7\% in YCbCr) \cite{ROSE}. However, the proposed scheme, which is also related to the LBP, achieves 10.9\% (HSV) and 11.9\% (YCbCr). Furthermore, from Table~\ref{tab:exp-both}, the proposed scheme achieves 0\% (YCbCr) on the IDIAP REPLAY-ATTACK database, which is better than the results of all the presented CNN-based methods and CoALBP. Therefore, it is concluded that the proposed scheme has achieved comparable performance to the state-of-the-art CNN methods and traditional methods.
	
	\begin{table}[tbp]
		\footnotesize
		\begin{center}
			\caption{Performance in terms of EER (\%) and HTER (\%) on MSU USSA database. The reuslts are obtained according to the 5-fold validation protocol in \cite{patel2016secure}.} \label{tab:exp-msu}%
			
			\begin{tabular}{|l|c|c|}
				\hline
				Method & EER & HTER \\
				\hline
				Patel et al. \cite{patel2016secure}& 3.84 & - \\
				\hline
				Patch-based CNN \cite{Atoum2018Face} & 0.55$\pm$0.26 & 0.41$\pm$0.32 \\
				\hline
				Depth-based CNN \cite{Atoum2018Face} & 2.62$\pm$0.73 & 2.22$\pm$0.66 \\
				\hline
				proposed (HSV) & 2.14$\pm$0.58 & 1.98$\pm$0.58 \\
				\hline
				proposed (YCbCr) & 1.56$\pm$0.61 & 1.33$\pm$0.51 \\
				\hline
			\end{tabular}%
		\end{center}
		\setlength{\textfloatsep}{0.1cm}
	\end{table}%

	\begin{table}[tbp]
		\footnotesize
		\begin{center}
			\caption{Performance (EER \%) of different numbers of trees in each forest.} \label{tab:exp-number_of_trees}%
			
			\begin{tabular}{|l|c|c|c|c|c|}
				\hline
				Dataset & Number of trees & 64 & 128 & 256 & 500  \\
				\hline
				\multirow{2}*{CASIA}   & HSV    & 8.62	&8.59&	8.67&	8.73     \\  \cline{2-6}
				  	&  YCbCr &9.53 &	9.54	&9.61&	9.66 \\ \hline
				
			    \multirow{2}*{IDIAP}   & HSV   & 0.054&	0.047&	0.048&	0.052\\ \cline{2-6}
			      &  YCbCr & 0.026 &0.017 &0.023 & 0 \\ \hline
			
				\multirow{2}*{MSU}   & HSV   & 1.99&	1.96&	2.22&	2.14\\  \cline{2-6}
			   	  &  YCbCr   & 1.26&	1.28&	1.42&	1.51\\ \hline
			   	 
				\multirow{2}*{ROSE-YOUTU}   & HSV  &  10.4 & 10.4& 10.7 &10.9    \\ \cline{2-6}
		    	 	&  YCbCr & 11.4& 11.5 &11.3& 11.9 \\
		    	 \hline

			\end{tabular}%
		\end{center}
		\setlength{\textfloatsep}{0.1cm}
	\end{table}%

	\par The experimental results on the MSU USSA database, in terms of the EER and the Half Total Error Rate (HTER), are provided in Table~\ref{tab:exp-msu}. According to Table~\ref{tab:exp-msu}, the patch-based CNN \cite{Atoum2018Face} achieves the best results both in EER (0.55\%) and HTER (0.41\%) on the MSU database, but our proposed scheme achieves 2.14\% EER and 1.98 \% HTER in HSV space as well as 1.56\% EER and 1.33\% HTER in YCbCr, which are better than that of the Depth-based CNN \cite{Atoum2018Face} with 2.62\% EER and 2.22\% HTER.
	\par In summary, taking Tables~\ref{tab:exp-both} and \ref{tab:exp-msu} together, our proposed method is highly competitive when compared with the state-of-the-art CNN-based methods and the traditional methods, \eg, CoALBP \cite{Nosaka2011Feature}.
	\subsubsection{Comparisons of different numbers of trees in each forest}
	\par In the above experiments, we follow \cite{gcforest} and adopt 500 trees in each forest. In a certain range, the more trees in a forest, the better the performance. However, too many trees in a forest would introduce heavy computational costs. In \cite{howmanytrees}, it is suggested that a trade-off between performance and computational costs can be achieved when the number of trees in a forest is in the range from 64 to 128. There are no significant performance gains when the number of trees increases to 512, 1024, 2048 or other larger numbers. Experimental results in Table 4 show that when the number of trees is smaller than 500, the performance does not necessarily drop. This observation coincides with the conclusion in \cite{howmanytrees}.

	\section{Conclusion and Future Work}\label{sec:end}
	\par Given the concern on the adversarial attack, in this paper, we propose to utilize the deep forest \cite{gcforest} in the problem of the FAS. To the best of our knowledge, this is the first attempt to introduce the deep forest into the FAS problem. Inspired by works related to texture analysis, we re-devise the constructing of multi-scale representations by integrating LBP descriptors with the deep forest learning scheme. Our proposed scheme has achieved better results than the original GSM proposed by \cite{gcforest}. Furthermore, compared with the state-of-the-art approaches, competitive results have been achieved on several benchmark databases by the proposed scheme. For example, 0\% EER is achieved on the IDIAP dataset. This indicates the effectiveness and competitiveness of our proposed scheme. Hence, our method could offer a competitive option to those who would like to improve the security of their systems by fusing diverse approaches in their schemes in system-level. Moreover, there have been a limited number of research works which exploit the deep forest on practical problems. This paper could serve as an important reference to the researchers who want to explore methods beyond the CNN-based schemes.
	\par Admittedly, the results of our approach do not look as attractive as some CNN-based methods. In the future, various efforts can be made to improve the overall performance, such as investigating more cascading strategies and feature extraction methods. In this work, the LBP is utilized because it is common in the field of the FAS problem and it is relatively simple for us to implement with the deep forest. However, the LBP is designed by the researchers in computer vision society based on their domain knowledge. Such knowledge may not be fully applicable to the FAS problem. Some novel methods of binary descriptors have raised our strong interest and given us significant references, \cite{SensitiveLBP_TIP2015, DeepHash_TIP2017, LBF_PAMI2018, ContextLBP_PAMI2018}. Designed in a more intellectual idea, they can learn features from data and are less dependent on people's knowledge. Hopefully, we could achieve better results by referring to these methods.


	\bibliographystyle{ieeetr}
	\bibliography{main}
	
	
\end{document}